\documentclass[letterpaper, 10 pt, conference]{ieeeconf}

\IEEEoverridecommandlockouts

\usepackage{blindtext, graphicx}
\usepackage[noend]{algpseudocode}
\usepackage{algorithm}
\usepackage{nicefrac}
\usepackage{amsfonts}
\usepackage{amsmath}
\usepackage{mathtools,nccmath}
\usepackage{tabularx}

\usepackage{etoolbox,siunitx}
\usepackage{multirow}
\usepackage{framed}
\usepackage{arydshln}
\usepackage{color}
\usepackage{tabularx}
\usepackage[dvipsnames]{xcolor}

\usepackage{hyperref}

\usepackage{caption}
\usepackage[percent]{overpic} 

\def \path {\mathbf{P}}

\newcommand{\etal}{\textit{et al}.}
\newcommand{\ie}{\textit{i}.\textit{e}. }
\newcommand{\eg}{\textit{e}.\textit{g}. }

\newcommand{\RNum}[1]{\uppercase\expandafter{\romannumeral #1\relax}}

\newcommand{\xxnote}[3]{}
\ifx\hidenotes\undefined
  \usepackage{color}
  \renewcommand{\xxnote}[3]{\color{#2}{#1: #3}}
\fi

\begin{document}

\title{\LARGE \bf Auto-conditioned Recurrent Mixture Density Networks \\ for Learning Generalizable Robot Skills}
\author{\renewcommand{\thefootnote}{\fnsymbol{footnote}}
Hejia Zhang, Eric Heiden, Stefanos Nikolaidis, Joseph J. Lim, Gaurav S. Sukhatme%
\thanks{Hejia Zhang, Eric Heiden, Stefanos Nikolaidis, Joseph J. Lim, and Gaurav S. Sukhatme are with the Department of 
Computer Science, University of Southern California, Los Angeles, USA 
{\texttt{\{hejiazha, heiden, nikolaid, limjj, gaurav\}@usc.edu}}.}%
}

\maketitle

\begin{abstract}
Personal robots assisting humans must perform complex manipulation tasks that are typically difficult to specify in traditional motion planning pipelines, where multiple objectives must be met and the high-level context be taken into consideration. Learning from demonstration (LfD) provides a promising way to learn these kind of complex manipulation skills even from non-technical users. However, it is challenging for existing LfD methods to efficiently learn skills that can generalize to task specifications that are not covered by demonstrations. In this paper, we introduce a state transition model (STM) that generates joint-space trajectories by imitating motions from expert behavior. Given a few demonstrations, we show in real robot experiments that the learned STM can quickly generalize to unseen tasks and synthesize motions having longer time horizons than the expert trajectories. Compared to conventional motion planners, our approach enables the robot to accomplish complex behaviors from high-level instructions without laborious hand-engineering of planning objectives, while being able to adapt to changing goals during the skill execution. In conjunction with a trajectory optimizer, our STM can construct a high-quality skeleton of a trajectory that can be further improved in smoothness and precision. In combination with a learned inverse dynamics model, we additionally present results where the STM is used as a high-level planner.
\end{abstract}

\IEEEpeerreviewmaketitle

\section{Introduction}
\label{sec:intro}

A promising direction toward the wide deployment of robots in human environments is in robotic personal assistants. To realize such goals where a robot can perform typical household chores, such as cooking or house cleaning, one of the roadblocks lies in attaining complex manipulation skills. Besides having a rich library of behaviors a robot can accomplish, it furthermore needs to be able to adapt these skills to individualized user preferences. 

In the study of robotic manipulation, it is often challenging to express complex manipulation problems in terms of a sequence of waypoints the end-effector should follow while opening and closing the gripper, as is commonly required by conventional motion planners. Given a typical task in a human environment, such as cleaning dishes, the precise description of the problem for the motion planner is difficult. On the other hand, \emph{learning from demonstration} (LfD) approaches use machine learning models to imitate expert behavior without a formal program that encodes the motion plan. In addition, tapping into the potential of deep learning models for motion planning has been reported to lead to two orders of magnitude in computation speed improvements over conventional planning algorithms~\cite{qureshi2018mpn}, such as \emph{Optimal Rapidly-exploring Random Trees} (RRT$^*$)~\cite{karaman2011rrtstar} and \emph{Batch Informed Trees} (BIT$^*$)~\cite{gammell2015bitstar}. LfD also provides the potential for non-technical users to teach robots new skills easily.

While numerous LfD algorithms have been proposed, \eg{}\emph{behavioral cloning}~\cite{atkeson1997robot}, \emph{inverse reinforcement learning}~\cite{ng2000algorithms} and \emph{generative adversarial imitation learning}~\cite{ho2016gail}, teaching robots generalizable skills is still challenging. We are particularly interested in the ability of skill models to perform tasks with unseen goals and plan tasks with longer time horizons than the demonstrated tasks. 

To address this gap, in this paper we propose a learned \emph{state transition model} (STM) that can imitate a variety of motions and map high-level task descriptions to state sequences. Such a model has the potential to generate trajectories on the basis of instructions that are intuitive to a non-technical human operator. Through our experiments, we show our proposed model has the generalizability that we described before.

Our work is motivated by recent advancements in computer graphics and robotics research. These work shows the potential of recurrent neural networks for synthesizing extended complex human motion sequences~\cite{zhou2018autoconditioned} and learning robot skills directly from expert demonstrations~\cite{rahmatizadeh2016lstmmdn}.


\begin{figure}
    \centering
    \includegraphics[width=.75\columnwidth]{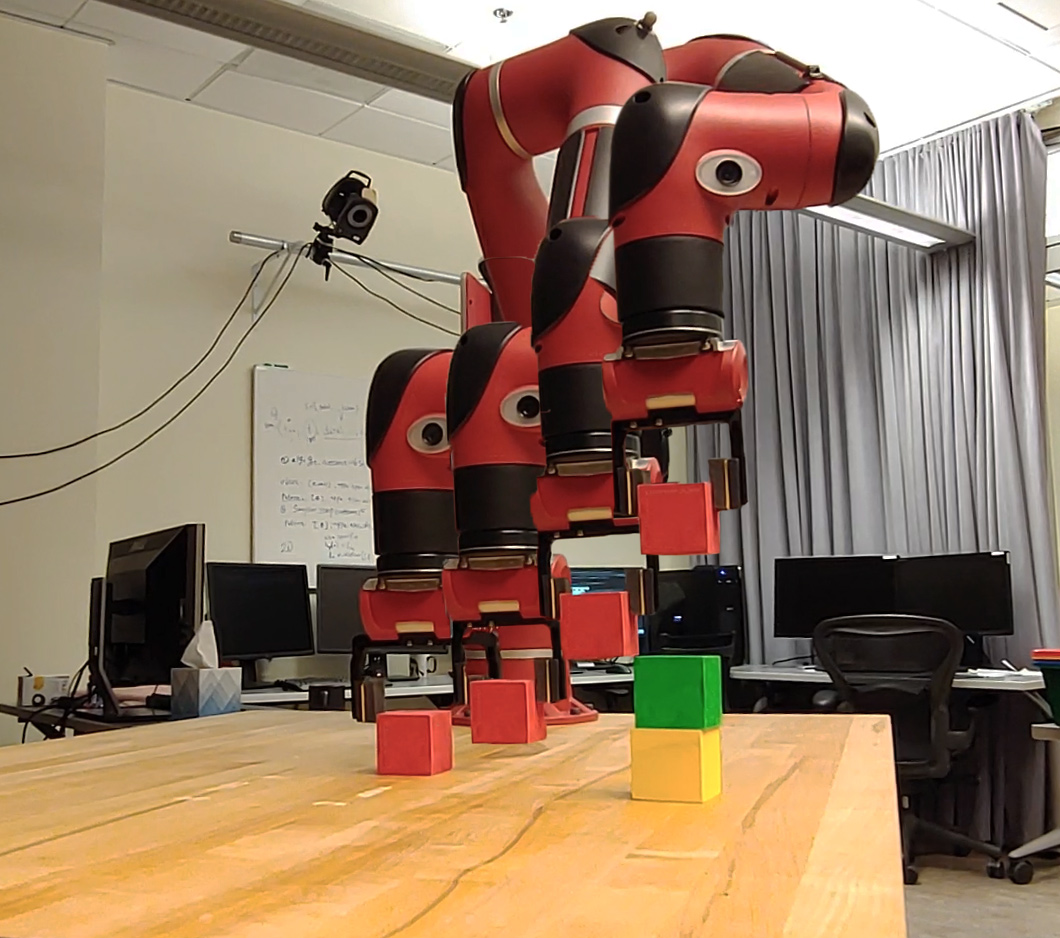}
    \caption{Action sequence of the block-stacking task on the Sawyer robot using the proposed state transition model (STM) synthesizing trajectories in joint-position space.}
    \label{fig:sawyer_stacking}
\end{figure}

\begin{figure*}
    \centering
    \includegraphics[width=.9\textwidth]{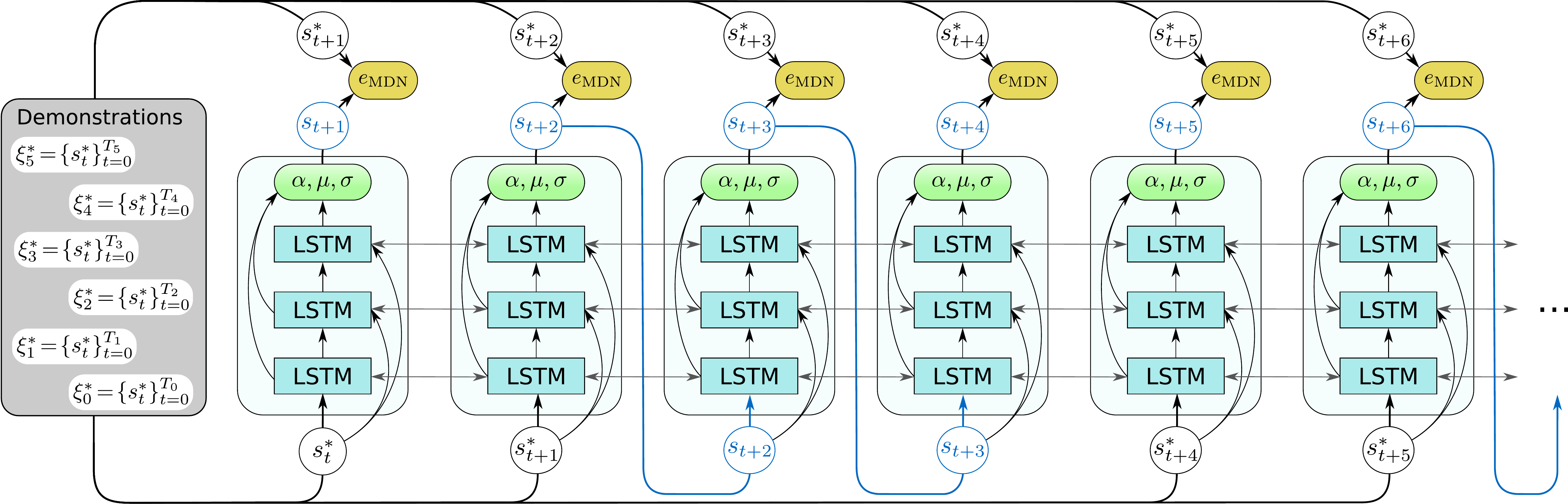}
    \caption{Architecture of the proposed auto-conditioned recurrent mixture density network to model state transitions, unrolled over 6 time steps, with an exemplary {\color{NavyBlue}{auto-conditioning length $v=2$}} and ground truth length $u=2$ (see Sec.~\ref{sec:lstm}).}
    \label{fig:architecture}
\end{figure*}


In this work, we build on \emph{recurrent neural networks} (RNN) as sequence learning models to learn a variety of robot motions. We demonstrate how a robot skill can be learned from a few demonstrations by leveraging a deep learning architecture and novel training methodology. As we show in our experiments, our model is able to synthesize circular trajectories with varying radii, generalizing from a sparse set of demonstration trajectories. We show that when we change the goal online during the trajectory generation, the model is able to adapt and replan accordingly for motions tasks that are longer than the demonstrated ones. In combination with trajectory optimization, we show the trajectory generated by our model can be easily optimized with subject to trajectory smoothness and task goals.

Our contributions are as follows:
\begin{enumerate}
    \item We present a training procedure and stochastic recurrent neural network architecture that can efficiently learn robot skills from demonstrations in joint position space.
    \item We provide real-robot experiments that demonstrate the ability of our STM to generalize to tasks that it has not been trained on. The generalizability allows our model to accomplish complex behaviors from high-level instructions which would traditionally require laborious hand-engineering and sequencing of trajectories from motion planners.
    \item We show our proposed model can be used to generate an initial trajectory skeleton which can be post-processed for different purposes: (1) In combination with trajectory optimizer, we show the initial trajectory can be further optimized to yield smooth trajectories that can accomplish goals more precisely; (2) In conjunction with a learned inverse dynamics model, we show results on transferring learned high-level skills in simulation to low-level control policies on the real robot.
\end{enumerate}

The paper is organized as follows: we first review related work in
Sec.~\ref{sec:related_work} and then formally define our problem in Sec.~\ref{sec:problem} and describe our deep learning architecture and training methodology in detail in Sec.~\ref{sec:our_approach}. Finally, through an experimental study, we evaluate how our learning framework can efficiently learn task-level generalizable robot skills for a seven degrees-of-freedom robot arm in Sec.~\ref{sec:experiments} and report results in Sec.~\ref{sec:results}.

\section{Related Work}
\label{sec:related_work}

Using machine learning models for motion planning and control is an active area in robotics research. One of the central robot learning paradigms is reinforcement learning~\cite{mnih2015humanlevel, duan2016rllab} that considers the robot interacting with its environment as a \emph{Markov decision process} (MDP) where the robot observes the environment, takes actions, and subsequently receives a reward signal. This formalism has been widely studied because of its benefit of learning from experience. However, since RL algorithms rely on sampling through extensive interaction with the environment, it is still challenging to deploy RL algorithms in the real world. Instead, in many current approaches, a transfer learning methodology is followed where a policy is trained in simulation and then transferred to the real world~\cite{christiano2016idm,james2017transfer,sadeghi2017transfer,julian2018scaling}. Similarly, in our paper, we present experimental results where our STM, which is trained in simulation, is used in conjunction with an inverse dynamics model that has been trained in the real world to transfer skills to the real world.

Another robot learning paradigm is \emph{learning from demonstration} (LfD)~\cite{atkeson1997robot, argall2009survey}, also referred to as imitation learning. In contrast to the model-free RL formalism, LfD is a sample-efficient robot learning method that leverages expert demonstrations and has been widely studied in the robotics research community. \emph{Behavioral cloning} approaches use supervised learning to train a model to imitate state-action sequences from an expert and have lead to early successes in autonomous driving~\cite{pomerleau1989alvinn}. However, naive behavioral cloning is prone to fail when there is a slight difference, also known as \emph{covariate shift}, between the demonstrated and the real-world experience~\cite{atkeson1997robot}.

To achieve more generalizable and robust robot skill learning, \emph{inverse reinforcement learning} (IRL)~\cite{ng2000algorithms} and \emph{apprenticeship learning} approaches~\cite{abbeel2004apprenticeship} attempt to recover the expert's reward function, given demonstrations from an expert policy which is assumed to optimize an unknown reward function $r^*\!\!:~\!S\to\mathbb{R}$, where $S$ denotes the set of states, such that a separate policy can be trained in a different context via reinforcement learning given that reward function. 

However, the learning pipeline of IRL is indirect and can be slow. Inspired by adversarial deep learning techniques for computer vision, such as generative adversarial networks (GAN), \emph{generative adversarial imitation learning} (GAIL)~\cite{ho2016gail, hausman2017multi, li2017infogail} approaches learn a policy, or generator, via reinforcement learning that aims to confound a separate discriminator network which classifies whether the policy's trajectory stemmed from the policy or from the expert.

In this paper, we study the problem of learning robot skills directly from a set of expert trajectories which are represented by sequences of states. Borrowing architectures and training methodologies from state-of-the-art sequence learning techniques~\cite{zhou2018autoconditioned}, our work addresses a fundamental issue in behavioral cloning which is the compounding error between the expert and the generated behavior over the course of the trajectory.

A special type of RNN, called the \emph{long short-term memory} (LSTM) cell~\cite{hochreiter1997lstm}, is widely used in time series prediction and sequence modelling, particularly in speech synthesis and speech recognition~\cite{wang2017autoregressive, graves2013speech}. Recent improvements, like \emph{auto-conditioning}~\cite{zhou2018autoconditioned} have shown great potential for synthesizing motions over hundreds of time steps, generating believable human movement patterns without drifting far from the motion capture demonstrations the LSTM was trained on. Leveraging such novel training methodologies, it becomes possible to generate trajectories over time spans longer than present in the training data. Similar approaches to auto-conditioning are \emph{professor forcing}~\cite{lamb2016professor}, \emph{data as demonstrator} (\textsc{DaD})~\cite{venkatraman2015dad} and \emph{dataset aggregation} (\textsc{DAgger})~\cite{ross2011dagger}. At scheduled intervals in the training procedure, these methods feed the RNN's previous outputs back into the RNN as input to the following cells to improve the prediction performance (\emph{cf}.~Fig.~\ref{fig:architecture}). Such training schedule mitigates drift from expert states while the RNN is unrolled over longer time spans without the input states that resemble the expert data.

A commonly used machine learning model to capture multimodal probability distributions is the \emph{mixture density network} (MDN)~\cite{bishop1994mdn} which represents multivariate Gaussian mixture models (GMM). Among other commonly used statistical models that can estimate probability distributions for estimation and control problems are kernel-based methods, such as \emph{Gaussian processes}~\cite{deisenroth2015gp}, and Monte-Carlo approaches, \eg{} \emph{particle filters}~\cite{thrun2002pf}.

Combining an RNN with an MDN has been first shown by Schuster~\cite{schuster1999mdnrnn} where the model is used to learn sequential data while capturing its stochasticity.
Similar to Rahmatizadeh \etal~\cite{rahmatizadeh2016lstmmdn, rahmatizadeh2018virtual}, we combine an LSTM with an MDN to architect the state transition model. In their work, the authors presented the feasibility of learning complex manipulation skills from imperfect demonstrations using the proposed method. In this paper, we focus on learning skills that are described by high-level task specifications, such as drawing a circle with the end-effector given a preset radius. We also perform the trajectory synthesis in the higher-dimensional joint position space, in contrast to Cartesian space, to remove the need for an inverse kinematics solver. Such an approach has the potential to lead to smoother trajectories without risking kinematic singularities. Moreover, thanks to auto-conditioning, our method can generate trajectories from a few demonstrations since in our training procedure the STM automatically learns to correct from states deviating from the demonstrations, whereas the method presented in~\cite{rahmatizadeh2016lstmmdn} uses explicit demonstrations that recover from undesired states back to the desired motion.

\section{Problem Definition}
\label{sec:problem}


Given $n$ expert trajectories $\{\xi^*_i\}_{i=0}^n$, where each trajectory $\xi^*_i$ is a state sequence $\{\mathbf{s}^*_{t_i}\}_{t_i=0}^{T_i}$ of length $T_i$, the problem is to estimate a model $p_\theta(\mathbf{s}_{t+1}|~\mathbf{s}_t)$ that, when unrolled for $T_i$ time steps from a start state $\mathbf{s}_0$, computes trajectories that resemble the expert demonstrations.

Throughout this work, we define a \emph{state} at a discrete time step $t$ as a vector of real numbers
\begin{align}
\label{eq:state}
\mathbf{s}_t = (\Delta q^0_t,\Delta q^1_t,\dots,\Delta q^6_t, \mathbf{\phi}_t, \mathbf{\psi}_t),
\end{align}
where $\Delta\mathbf{q}=\{\Delta q^j_t\}_{j=0}^6$ describes the changes in joint angles relative to the previous time step, $\mathbf{\phi}_t$ and $\mathbf{\psi}_t$ are vectors that denote the \emph{task-specific input} and the \emph{task description}, respectively. The latter two parameters vary in dimensionality, depending on the skill the robot is learning, and are detailed for each particular experiment in Sec.~\ref{sec:experiments}.


\section{Methodology}
\label{sec:our_approach}

The STM $p_\theta (\mathbf{s}_{t+1}|~\mathbf{s}_t)$ is a machine learning model parameterized by vector $\theta$ that captures the probability distribution over state transitions between the current state $\mathbf{s}_t$ and the next state $\mathbf{s}_{t+1}$. We select the model based on two important properties: (1) representing uncertainty in the state transitions and (2) being able to remember long sequences of states.

\subsection{Mixture Density Network (MDN)}
\label{sec:mdn}
Capturing the stochasticity of the state transitions is an integral ingredient for the deployment of our model on a real robot as future states are uncertain and high-dimensional. To address our first requirement of representing uncertainty, we use a \emph{mixture density network} (MDN)~\cite{bishop1994mdn} to estimate the probability distribution of future states.

The MDN parameterizes a multivariate mixture of Gaussians by estimating the distribution over the next states as a linear combination of Gaussian kernels:
$$
p(\mathbf{s}_{t+1}|\mathbf{s}_t) = \sum_{i=1}^m \alpha_i(\mathbf{s}_t) g_i(\mathbf{s}_{t+1} | \mathbf{s}_t),
$$

where $m$ is the number of Gaussians modelled by the MDN, $\alpha_i$ is the learned mixing coefficient and $g_i(\mathbf{s}_{t+1} | \mathbf{s}_t)$ is the $i$-th Gaussian kernel of the form
$$
g(\mathbf{s}_{t+1} | \mathbf{s}_t) = \frac{1}{\sqrt{2\pi} \sigma_i(\mathbf{s}_t)}\exp\left\lbrace -\frac{||\mathbf{s}_{t+1} - \mathbf{\mu}_i(\mathbf{s}_t)||^2}{2\sigma_i(\mathbf{s}_t)^2} \right\rbrace.
$$

In addition to $\alpha_i$, the kernel mean $\mu_i$ and standard deviation $\sigma_i$ are learned by the MDN.

Given the ground-truth state pair $(\mathbf{s}^*_t, \mathbf{s}^*_{t+1})$, we define the MDN loss as the negative log-likelihood:
\begin{align}
\label{eq:mdn_loss}
e_{\text{MDN}} = -\ln\left\lbrace \sum_{i=1}^m \alpha_i(\mathbf{s}^*_t) g_i(\mathbf{s}^*_{t+1} | \mathbf{s}^*_t) \right\rbrace.
\end{align}

\subsection{Long Short-Term Memory (LSTM)}
\label{sec:lstm}
To learn sequences of states, we require a model with an internal memory that allows it to remember states over long time horizons. As shown in Fig.~\ref{fig:architecture}, we propose to use the \emph{long short-term memory} (LSTM)~\cite{hochreiter1997lstm} architecture that maintains a hidden state $\mathbf{h}_t$. This allows the STM to make predictions of states over long time horizons. Therefore, the prediction $\mathbf{s}_{t+1}$ of our state transition model not only depends directly on the current state $\mathbf{s}_{t}$ but also on the hidden states of the LSTM cells.


We train the recurrent MDN with auto-conditioning~\cite{li2017sequential}, a learning schedule that, for every $u$ iterations of a sequence of $v$ time steps, feeds the LSTM's output as input into the cell computing the next state~(Fig.~\ref{fig:architecture}). This enables the network to correct itself from states that deviate from demonstrations: by learning from inputs where the network diverges from expert behavior, we capture the distribution of inputs that would cause a compounding error when rolling out the STM in the real world, where the expert demonstrations $\xi_i^*$ are no longer available as inputs to the network. This technique greatly improves the performance, as we report in our ablation study in Sec.~\ref{sec:results}.

Unlike the auto-conditioned LSTM from~\cite{zhou2018autoconditioned}, state vector $\psi_t$ representing the task description is designed to be modifiable from outside sources, such as a human operator, at every time step. Instead of only predicting one trajectory that is followed in an open-loop control fashion, in our framework the goal can be changed during the execution and the STM is able to replan the trajectory.

The overall loss function of our model is computed for each given expert trajectory $\xi^*_i$ by feeding the expert states and the model's own predictions according to the auto-conditioning schedule. By evaluating the recurrent MDN at every time step $t= 0,...,T_i$, we retrieve a trajectory of MDN outputs $\alpha_t, \mu_t, \Sigma_t$ and compute the MDN loss from~\eqref{eq:mdn_loss}. We implement our model in the automatic differentiation framework PyTorch that allows us to compute the gradients of the parameters $\theta$ of our model, for the recurrent and the MDN components, with respect to the input states. By \emph{backpropagating-through-time} (BPTT) the gradients, we update $\theta$ to minimize the loss via the Adam optimizer~\cite{kingma2014adam}.

\section{Experiments}
\label{sec:experiments}

In our experiments we focus on real-robot applications of our proposed STM architecture and training procedure, and demonstrate various use cases of the proposed model. We rely on inverse kinematics (IK) solvers, traditional motion planners and simulators to collect expert demonstrations and train the STM for the Sawyer robot, a seven-degrees-of-freedom robot arm, equipped with a parallel gripper as end-effector (\emph{cf.} Fig. \ref{fig:sawyer_stacking}).

We collected demonstration trajectories, \ie sequences of states $\lbrace s_t\rbrace_{t=0}^T$ of varying lengths in the Gazebo simulator~\cite{koenig2004gazebo} by using the IK solver provided by Rethink Robotics for Sawyer. On the real robot, the synthesized trajectories are executed using the joint-position control mode by forward integrating the first seven joint angle changes $\Delta\mathbf{q}_t$ from each state (\emph{cf.}~Eq.~\eqref{eq:state}).

\subsection{Sawyer Reacher}
\label{sec:reacher}
\begin{figure}
    \centering
    \includegraphics[height=3.15cm]{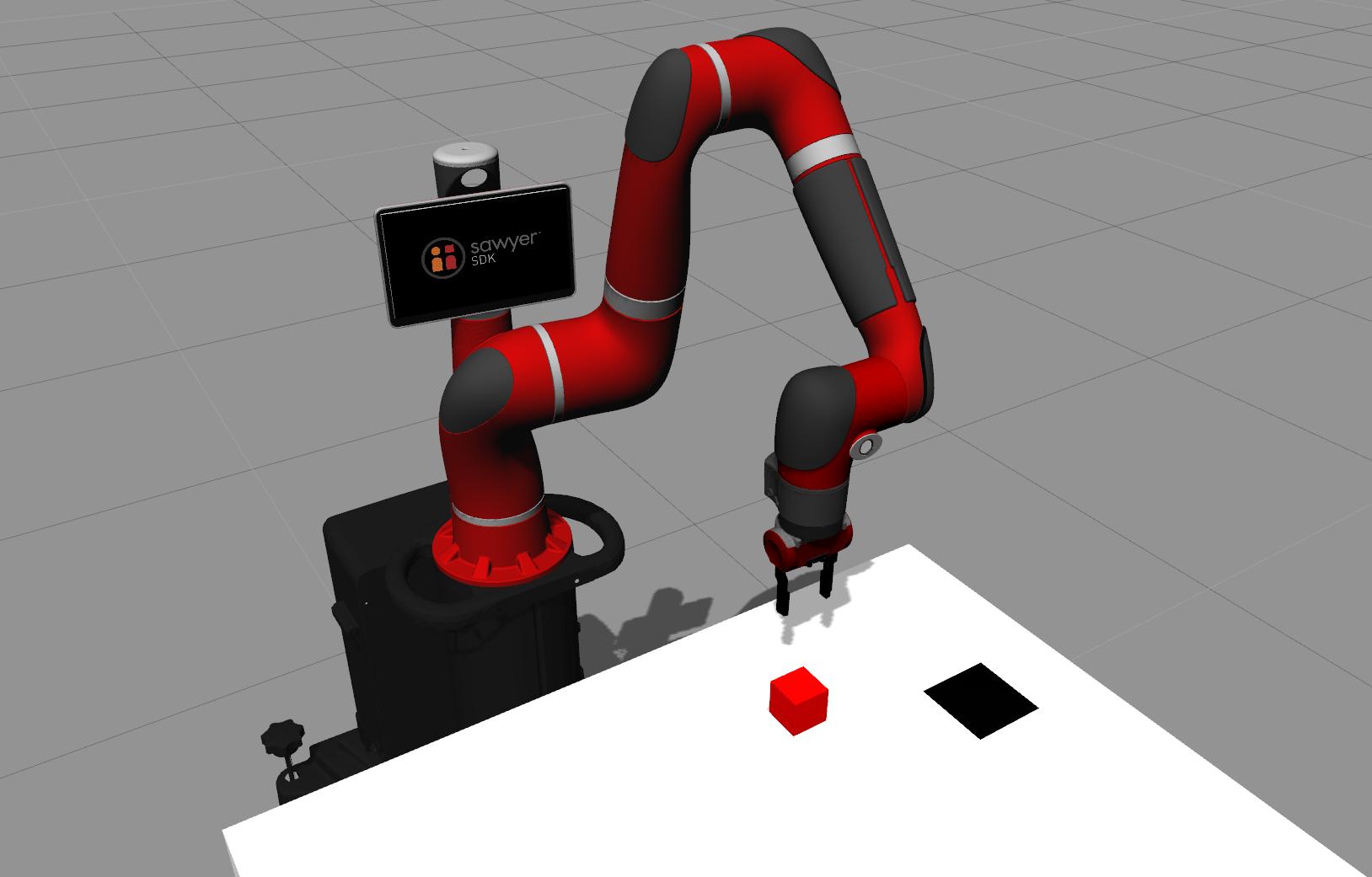}\hfill
    \includegraphics[trim=15cm 0cm 0cm 0cm,clip,height=3.15cm]{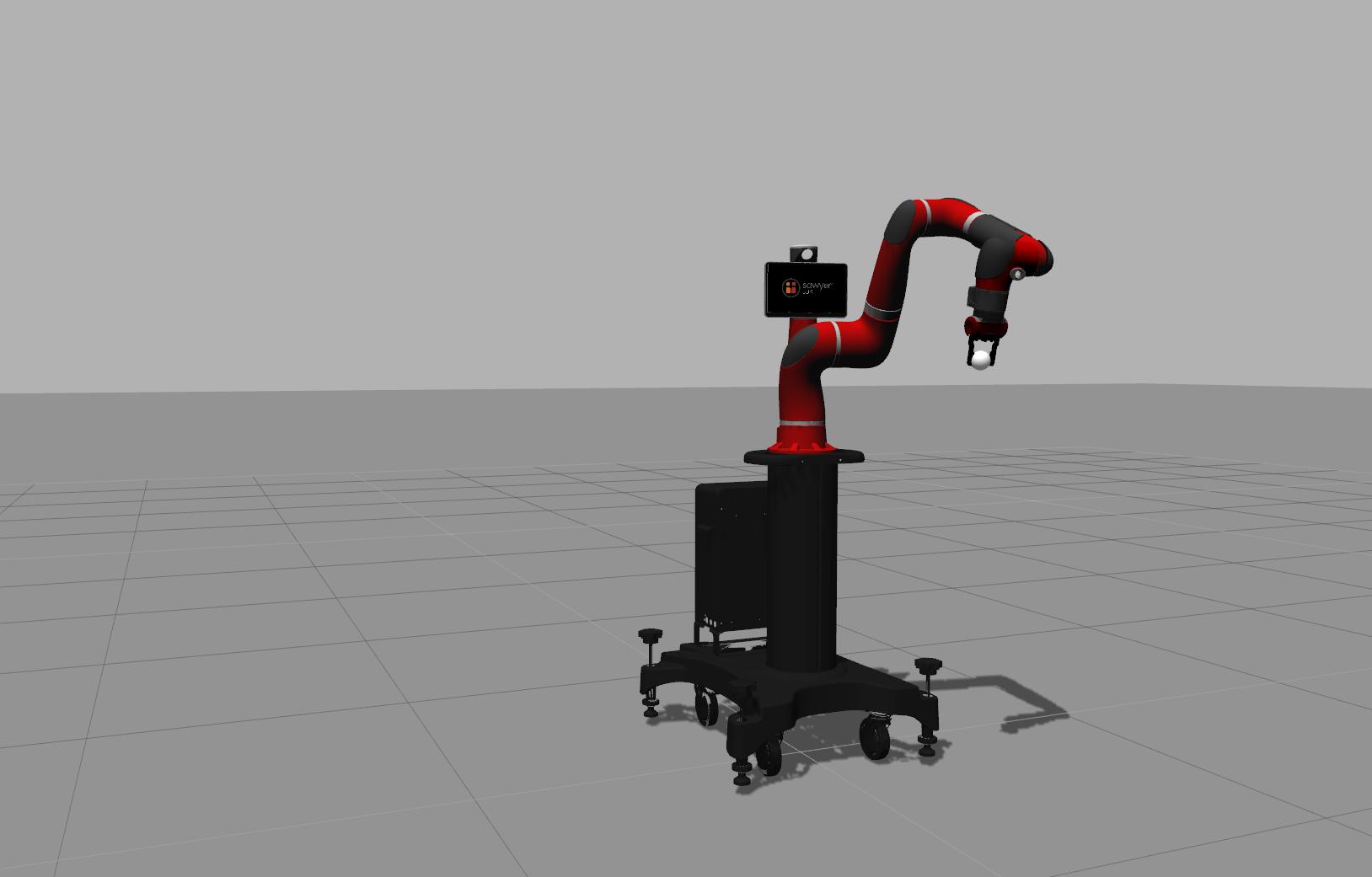}
    \caption{\emph{Left}: experimental setup for the Sawyer pick-and-place and block-stacking task in the Gazebo simulator. For pick-and-place, the robot is tasked to grasp a block from a random location and place it to a designated goal location in Cartesian space. In the block-stacking task, a tower of three blocks has to be built. \emph{Right:} simulation environment for the Sawyer reaching task in Gazebo. The objective for the STM is to synthesize a sequence of joint positions that servo the gripper from a random start configuration to a given goal location.}
    \label{fig:sawyer_pnp_setup}
\end{figure}

In the first experiment, we evaluate the STM on a basic servoing task: the STM is used to synthesize state sequences that move the gripper from a random initial joint configuration to a randomly sampled goal position (Fig.~\ref{fig:sawyer_pnp_setup}). The task-specific input $\phi_t$ is the three-dimensional end-effector position relative to the goal, and the task description $\psi_t$ is defined by the Cartesian coordinates of the goal location. In simulation, we collect 45 demonstration trajectories ranging between 50 and 70 time steps using the IK solver.

\subsection{Sawyer Pick-and-Place and Block Stacking}
\label{sec:stacking}
\begin{figure}
    \centering
    \includegraphics[width=1\columnwidth]{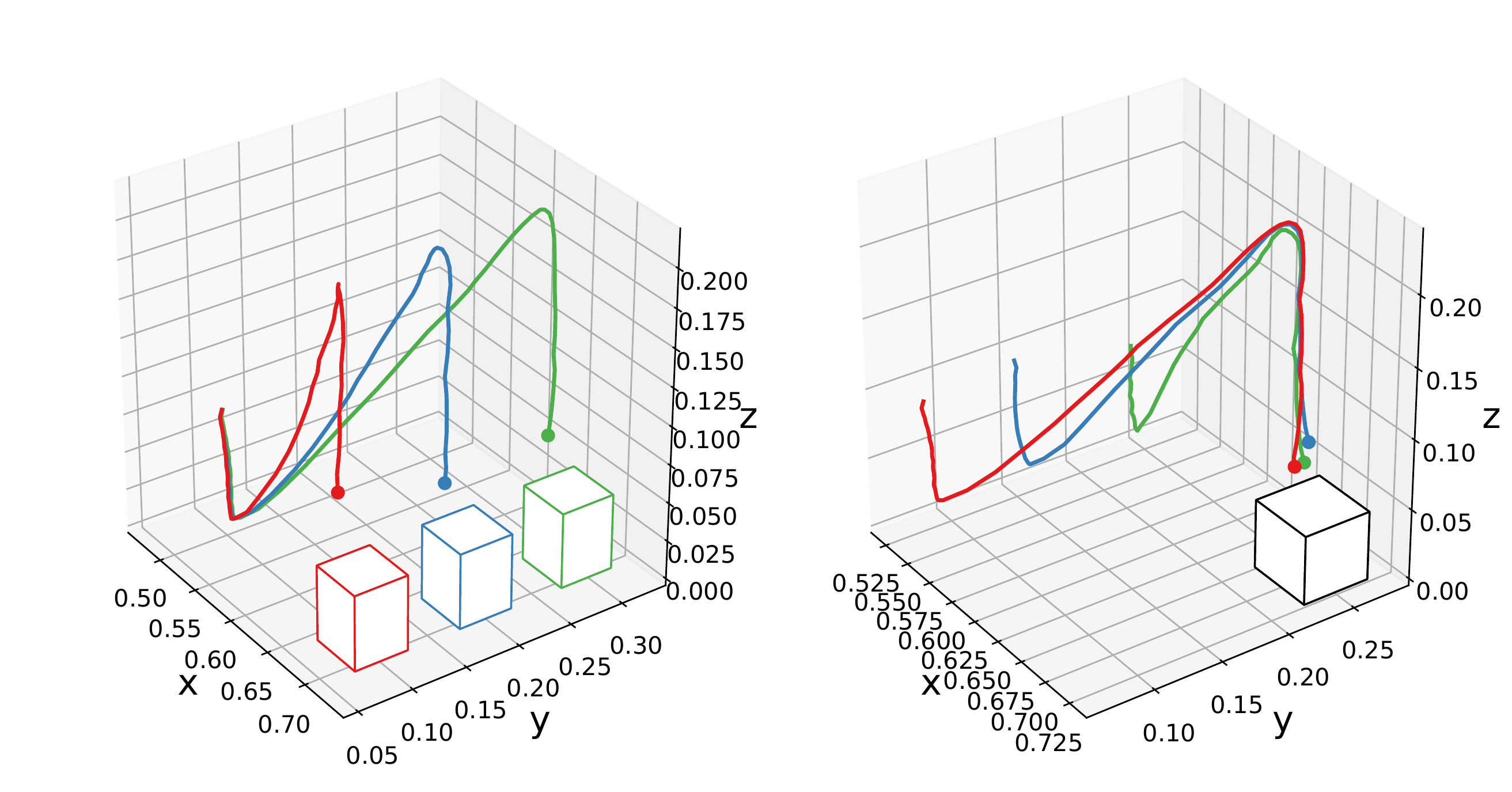}
    \caption{\emph{Left:} Gripper position trajectories of the three pick-and-place tasks where the robot is tasked to start from a fixed position, pick up the box and place it at three different goal positions. \emph{Right:} Trajectory of the gripper over the three pick-and-place tasks where the robot is tasked to start from three different configurations and place the block at a given goal position. The white blocks represent the task goals.}
    \label{fig:sawyer_pnp_gripperpos}
\end{figure}

In the second experiment, we evaluate the STM on pick-and-place and block-stacking tasks (Figs.~\ref{fig:sawyer_stacking},~\ref{fig:sawyer_pnp_setup}). We only need to train a single model for both tasks, which demonstrates the generalizability of our approach. In the block-stacking task, the STM needs to learn to stack three blocks at a random position with high precision. The STM is initialized by random initial joint configurations, while the blocks are placed in the same initial position at all times. The goal position of the blocks is randomly chosen. The task-specific input includes the three-dimensional end-effector positions $\phi_t$ relative to the goal and the Cartesian coordinates $\psi_t$ of the goal. We collect 150 demonstration trajectories in simulation, ranging from 166 to 170 states.

\subsection{High-level Control}
\label{sec:hlc}


In the next experiment, we evaluate how well our model can be used to perform tasks where only high-level task descriptions are given. We ask the robot to draw a circle of a defined radius $r$ and train the STM from a set of $10$ circular motion sequences as demonstrations, ranging from circles of radii between \SI{5}{\cm} and \SI{20}{\cm}. The task is described solely by $\psi_t = r$ and the task-specific input $\phi_t$ is given as the three-dimensional end-effector location.

Achieving such behavior with a traditional motion planning setup requires defining the waypoints on the circle such that the IK solver can compute the joint angles to transition between them. Converting between Cartesian and joint-space coordinates is subject to risking kinematic singularities. Although, in this work, we rely on IK solvers as demonstration source, our model can leverage other expert demonstrations, such as reinforcement learning agents or humans. Furthermore, by learning from demonstrations, a deep learning model could identify a connection between high-level goals (\eg the given radius) and the desired low-level behavior (\eg circle-drawing trajectories).

\subsection{Adapting Online to Changing Goals}
\label{sec:adapt}
An interesting fact about the auto-conditioned LSTM is that it can generate motion sequences than are much longer than the demonstrated sequences. This property is particularly useful in long-horizon motion planning robotics applications that include tasks decomposable into subtasks. We investigate this property on a reaching task, where we change the target in the middle of trajectory rollout. To adapt to changing targets, our model has to plan motions that exceed the horizons of its demonstration data. The experiments are performed both in reaching and pick-and-place tasks. 

\subsection{Open-loop Control with Inverse Dynamics Model}
\label{sec:stm_idm}
\begin{figure}
    \centering
    \includegraphics[width=.5\columnwidth]{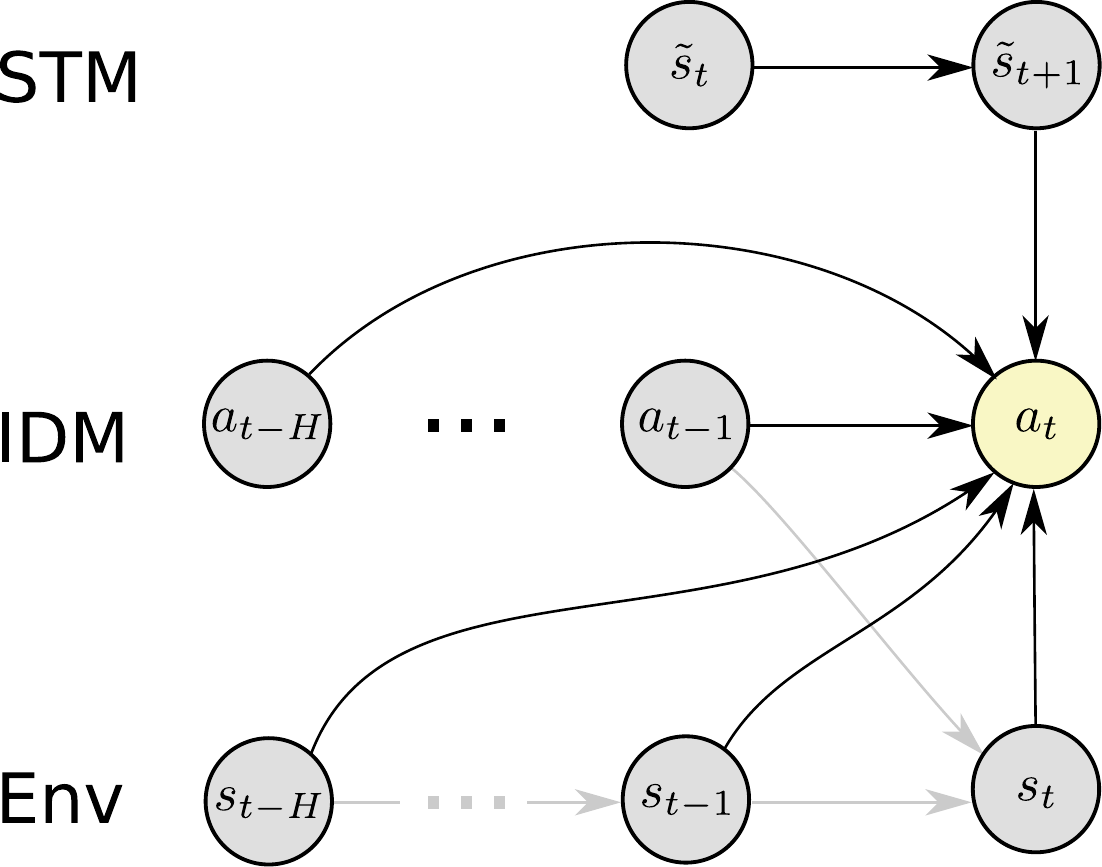}
    \caption{Graphical model of our open-loop control approach combining learned models for state transitions (STM) and inverse dynamics (IDM). Action $a_t$ is computed by the IDM given a history of the last $H$ steps~$\lbrace s_{i}\rbrace_{i=t-H}^t$ from the environment, the desired state~$\tilde{s}_{t+1}$ from the STM, and the previous actions~$\lbrace a_{i}\rbrace_{i=t-H}^{t-1}$.}
    \label{fig:graphical_model}
\end{figure}

\begin{figure*}
    \centering
    \includegraphics[width=.725\textwidth]{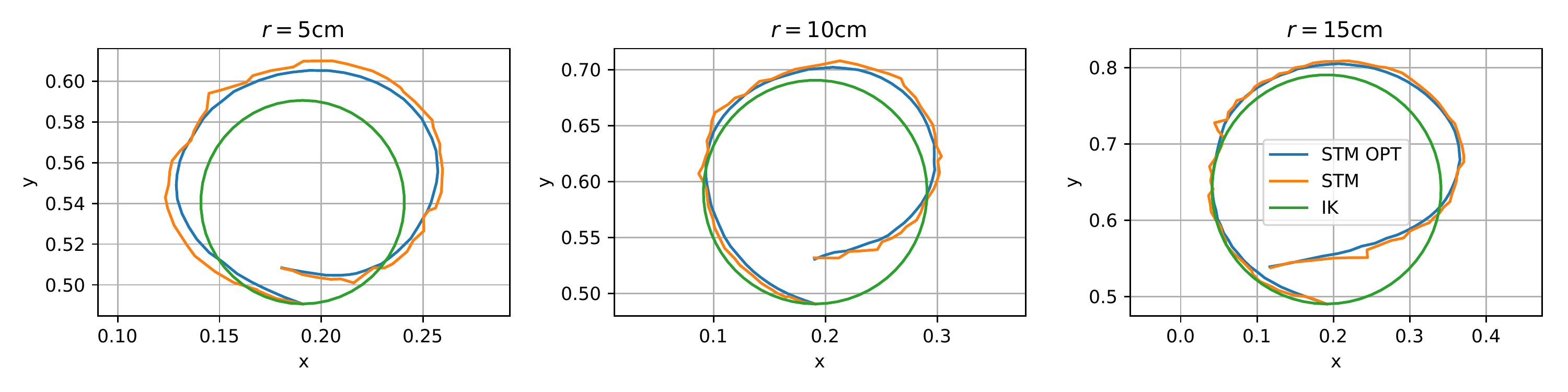}\hfill
    \includegraphics[width=.27\textwidth]{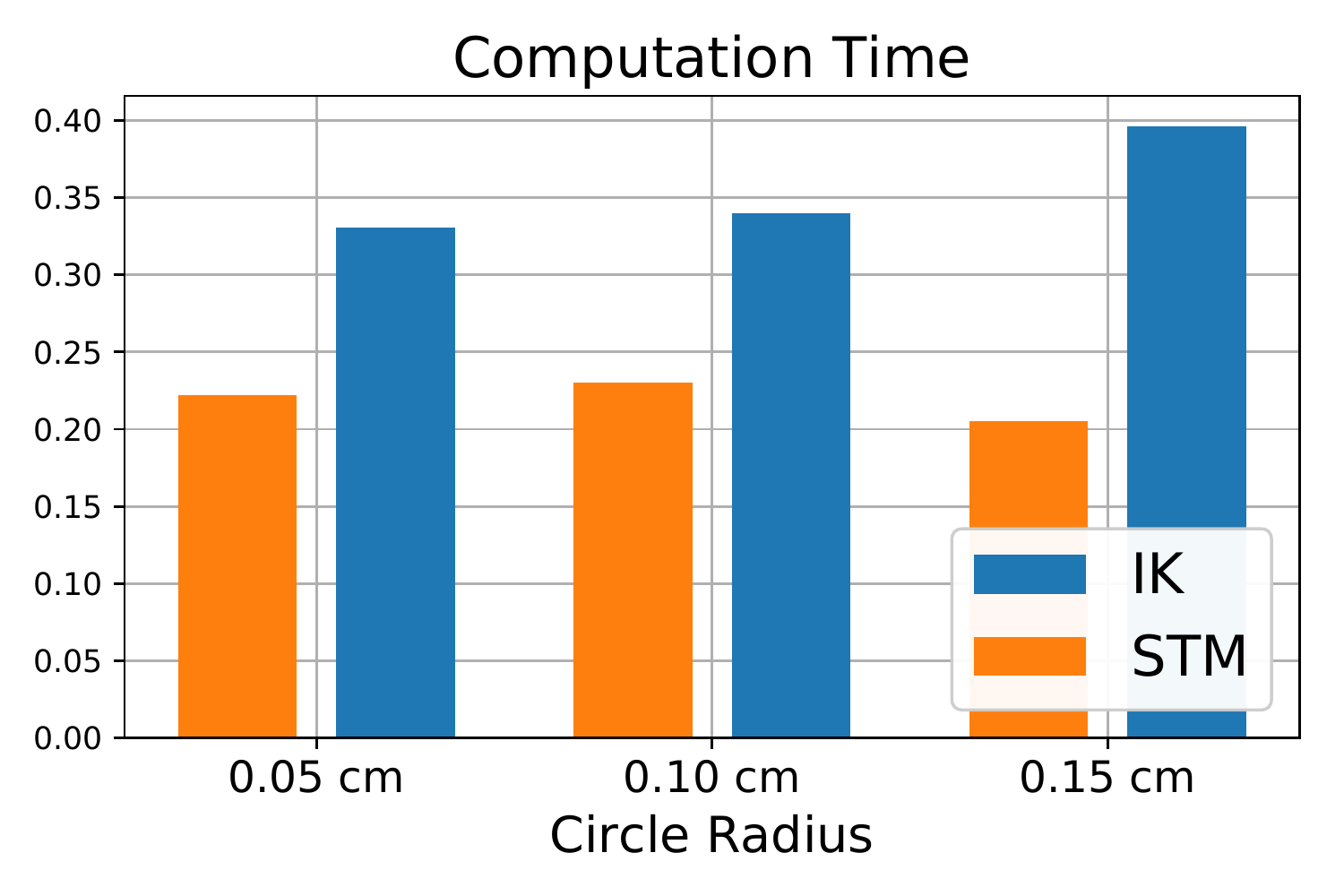}
    \caption{\emph{Left:} Gripper position trajectories for drawing circles of varying radii from the IK solver, the STM and the STM in combination with a trajectory optimizer. \emph{Right:} computation times for generating the circular trajectories using the STM and the IK solver.}
    \label{fig:draw_circle}
\end{figure*}

We trained an \emph{inverse dynamics model} (IDM) to accomplish torque control on the real robot. Combining an STM and IDM has the advantage of transferring from high-level behaviors in simulation to low-level control policies in reality: the STM, serving as joint position motion planner, remains unchanged between both environments. The IDM, on the other hand, can be trained separately on the real world, since it is the only module that depends on the environment dynamics. Such decoupling of both models has the potential for a higher sample efficiency compared to the simulation-to-real transfer of entire policy networks, as commonly done in traditional deep reinforcement learning approaches that train entirely in simulation~\cite{james2017transfer, sadeghi2017transfer, tan2018transfer}.

The IDM is a three-layer MDN (each layer having 256 hidden units) that parameterizes a Gaussian mixture model consisting of fifteen normal distributions per action dimension (seven dimensions for joint actuators). Through our experiments, we found an IDM conditioned on the current state, plus the two previous states and actions (\emph{cf}. Fig.~\ref{fig:graphical_model} for $H=2$), to achieve the highest accuracy in steering between $s_t$ and $s_{t+1}$ via torque control.

We investigate the combination of STM and IDM on a simple Sawyer reaching testbed where the robot is tasked to servo the gripper to one of four desired goal locations.
We train the IDM in Gazebo with the \emph{Open Dynamics Engine}~\cite{drumwright2010ode} physics simulation, and with the  \emph{Bullet}~\cite{coumans2015bullet} physics engine.

\subsection{STM as Initial Solution for Trajectory Optimization}
\label{sec:trajopt}
To combine data-driven methods with trajectory optimization methods, following the work by Kratzer~\etal~\cite{kratzer2018towards}, we sample from our model first to generate a feasible initial trajectory skeleton, $\lbrace \tilde{\mathbf{q}}_t \rbrace_{t=1}^T$. We want to retain the shape of the initial trajectory, while improving its smoothness. We can also encode task goals as the additional cost objectives, such as the Euclidean distance between the last state of the generated trajectory and the target position in the reaching task.

Similar to Kratzer \etal~\cite{kratzer2018towards}, the optimizer adapts the joint positions $\{\mathbf{q}_t\}_{t=0}^T$ of the trajectory generated by the STM by minimizing the objective

\begin{align}
\nonumber
\lbrace \mathbf{q}^*_t \rbrace_{t=1}^T &= \operatorname*{arg~min}_{\lbrace \mathbf{q}_t \rbrace_{t=1}^T} V(\lbrace \mathbf{q}_t \rbrace_{t=1}^T)
\intertext{where the cost function is}
\label{eq:traj_opt}
V(\lbrace \mathbf{q}_t \rbrace_{t=1}^T) &= \sum_{t=1}^{T-1} \| \mathbf{q}_{t} - \tilde{\mathbf{q}}_t \|_2^2 +  \gamma \| \mathbf{q}_{t+1} - \mathbf{q}_{t} \|_2^2.
\end{align}

At every iteration, the cost function trades off two opposing objectives, weighted by the coefficient $\gamma$. The term $\| \mathbf{q}_{t+1} - \mathbf{q}_{t} \|_2^2$ has a smoothing effect on the trajectory. When the start and goal state are kept constant, solely optimizing for this term would result in a straight line. To retain the shape of the original trajectory, the term $\| \mathbf{q}_{t} - \tilde{\mathbf{q}}_t \|_2^2$ ensures closeness between the current joint positions $\mathbf{q}_t$ on the smoothed trajectory and the joint positions $\tilde{\mathbf{q}}_t$ on the trajectory prior to the smoothing iteration. As in related trajectory optimization frameworks (\eg{} CHOMP \cite{ratliff2009chomp}), our cost function can be extended to account for obstacles and other objectives. We minimize the cost function using an iterative gradient descent method which, with appropriate tuning of the weighting coefficient $\gamma$, results in smooth trajectories that still resemble the original complex motions. 

\section{Results}
\label{sec:results}

\newcommand\Tstrut{\rule{0pt}{2.6ex}}         
\newcommand\Bstrut{\rule[-0.9ex]{0pt}{0pt}}   
\begin{table}[t]
    \centering
    \begin{tabularx}{\columnwidth}{X|ccc}
         &          \bf Reacher & \bf Pick-and-place & \bf Stacking \Tstrut\Bstrut \\ \hline
        \bf LSTM & 80\%       & 0\%               & 0\%        \Tstrut\Bstrut\\
        \bf a.c. LSTM & 90\%       &  50\%               & 25\%         \Bstrut\\
        \bf LSTM-MDN & 90\%       & 60\%               & 30\%         \Bstrut\\
        \bf a.c. LSTM-MDN (Ours) & 100\%       & 100\%               & 80\%         \\
    \end{tabularx}
    \caption{Success rates for the experiments described in Sec.~\ref{sec:experiments} over 20 roll-outs with varying architectures and training procedures. All models have been trained on $X$ demonstrations over $Y$ training iterations. The reaching task is successful if the gripper is within \SI{5}{\cm} of the goal position by the end of the trajectory. The STM's for reacher are evaluated in simulation, pick-and-place and stacking success rates come from real-robot experiments.}
    \label{tab:success_rates}
\end{table}

We wish to investigate the benefit of combining LSTM and MDN for skill learning. Therefore, we compare LSTM-MDN against two baselines, a plain LSTM, and an auto-conditioned LSTM. We present the results in Table~\ref{tab:success_rates}. We measure performance using the success rates over 20 rollouts for each of the experiments described in Sec.~\ref{sec:experiments}. Our method outperforms other baseline models on all of the tasks, especially on pick-and-place and block stacking, where more complex trajectories need to be synthesized.

\begin{figure}
    \centering
    \includegraphics[width=1\columnwidth]{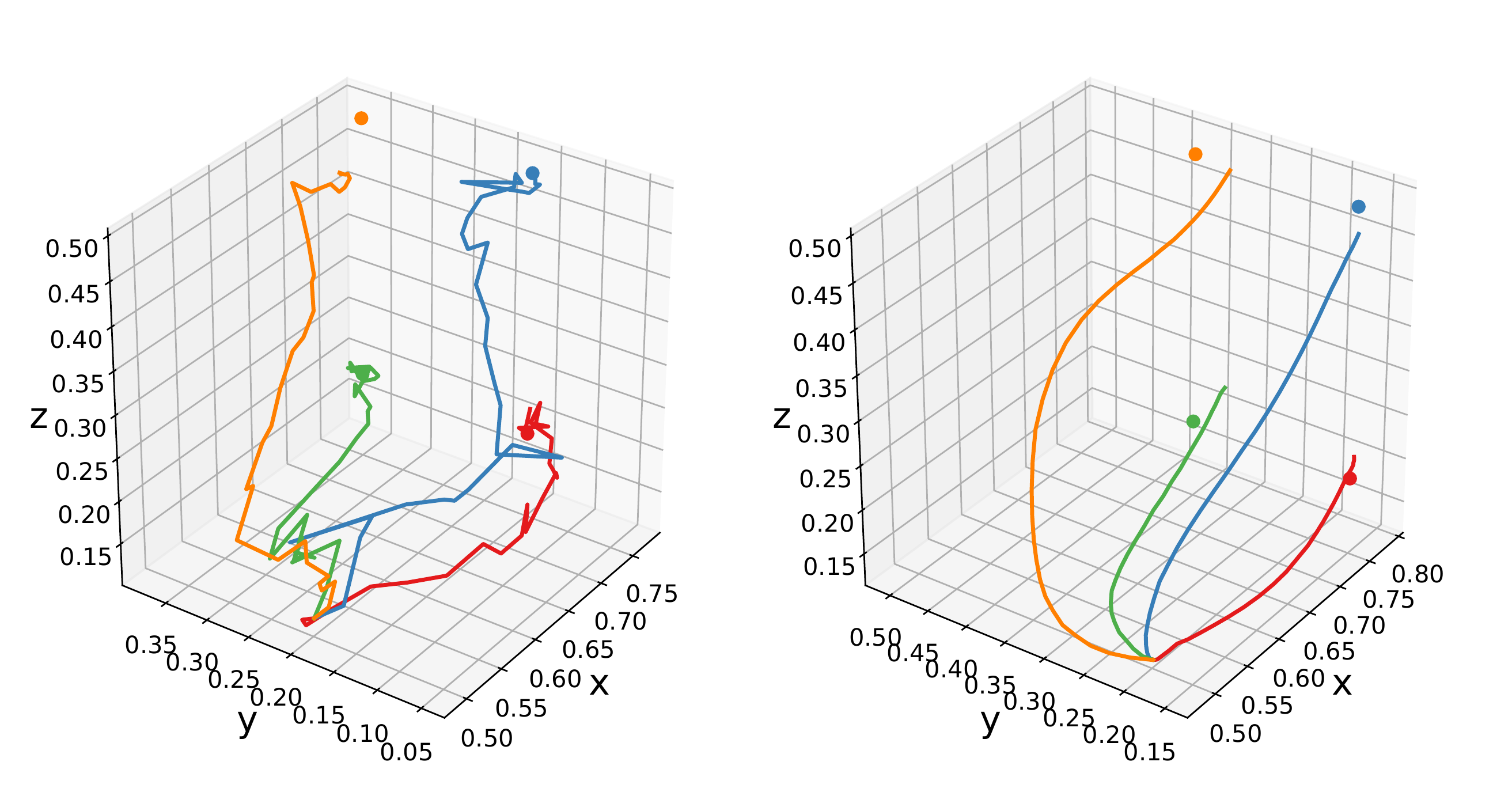}
    \caption{Before (left) and after (right) smoothed gripper position trajectories (lines) from forward kinematics rollouts on state sequences generated by the STM. The robot is tasked to servo its gripper to four goal positions (dots).}
    \label{fig:sawyer_reacher_gripperpos}
\end{figure}

All of our experiments, including model training and execution, are conducted on a personal computer equipped with an \textsc{Nvidia} GeForce GTX 1070 graphics card.

For the reaching task, our proposed STM trained over 5,000 training iterations in ca. \SI{10}{\min}. For comparison, we use a three-layer LSTM with 64 hidden units per layer for all of the models, and three Gaussians in the MDN-based STMs, \ie vanilla LSTM-MDN and auto-conditioned LSTM-MDN. We observed that despite the simplicity of the task, the MDN yielded more robust behavior than plain LSTM structure. The STM baseline without our stochastic model failed to find any trajectories that reached close to the goals.

For the pick-and-place and block-stacking tasks (see Fig.~\ref{fig:sawyer_pnp_gripperpos}), our proposed STM trained over 30,000 training iterations in ca. two hours. We use the three-layer LSTM with 128 hidden units per layer for all of the models and 20 Gaussians on the MDN-based models. We observed that auto-conditioning significantly reduces the accumulation of error, which is a common problem in generating trajectories using RNNs.

\begin{figure*}
    \centering
    \includegraphics[width=1.7\columnwidth]{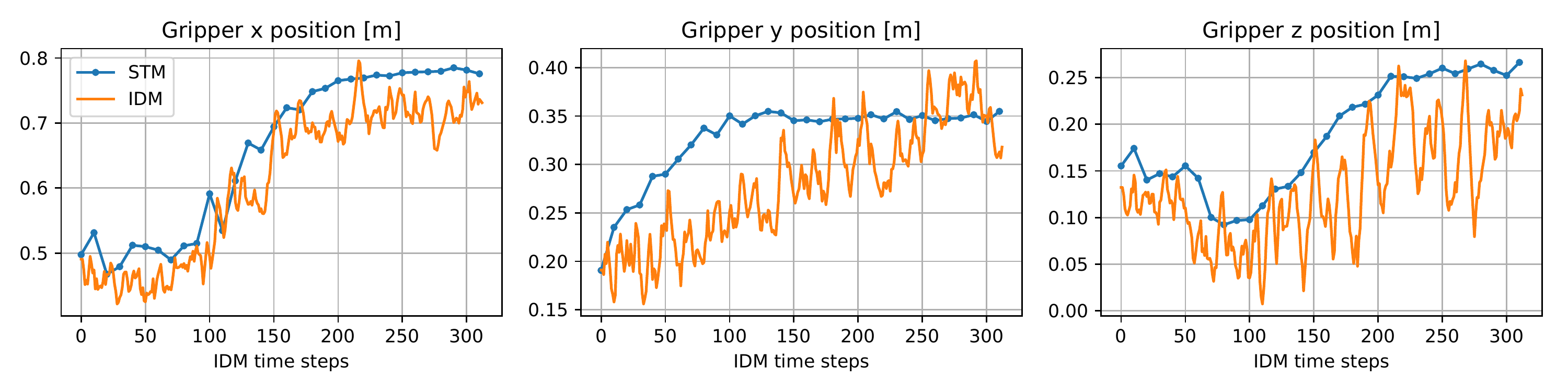}
    \caption{Tracking performance of the IDM to generate joint torque commands that follow the state sequence synthesized by the STM on the Sawyer reaching task. The IDM produces actions at a higher frequency (10 times as fast) than the STM. Shown are the Cartesian coordinates of the gripper where the joint angles from the STM are played back through forward kinematics and the IDM is deployed in Gazebo with the ODE physics engine.}
    \label{fig:reacher_tracking}
\end{figure*}

We also show the result trajectory of the system combining the STM and trajectory optimization methods, which shows that the proposed STM can generate an initial trajectory skeleton for trajectory optimization (Fig.~\ref{fig:draw_circle}). 

We additionally show the benefits of a deep learning model as high-level controller and investigate its ability of associating high-level commands with demonstration trajectories, as described in Sec.~\ref{sec:hlc}. As shown in Fig.~\ref{fig:draw_circle}, our STM is able to learn from a few demonstrations the connection between the radius and the resulting trajectory. The average root mean square errors (RMSE) of the generated trajectories for the three radii are 0.017, 0.028 and 0.110, respectively -- resulting in a circular motion that never exceeds an error of $10\%$ in radius at any point on the trajectory. Another benefit is in the computation times of the STM: evaluating the STM on the circle-drawing task is almost twice as fast as obtaining a solution from an IK solver (\emph{cf}. Fig.~\ref{fig:draw_circle}).

\begin{figure}
    \centering
    \includegraphics[width=1\columnwidth]{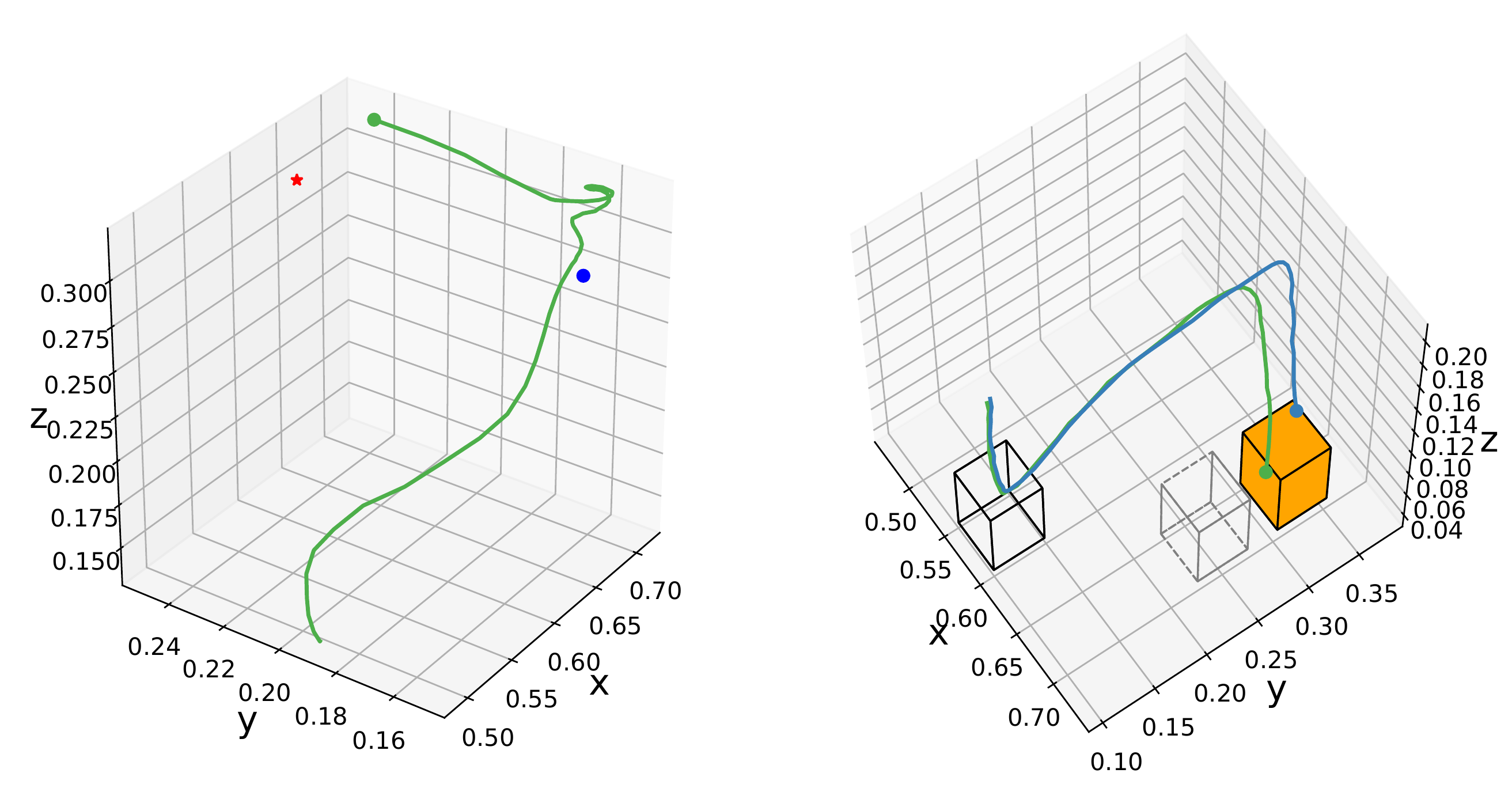}
    \caption{Plot of the gripper position trajectory generated by the STM for the goal changing reacher (Left) and pick-and-place (Right) task. For reacher task plot, blue dot is at the old goal position while red start is at the new goal position. For pick-and-place task plot, white block is at the start position, transparent block is the old goal while orange block is the new goal. The goal is changed at the middle of each task execution. The plot shows how our model can adapt to changing goal and still work beyond the planning horizon of its demonstrations.}
    \label{fig:sawyer_changing_goals}
\end{figure}

To evaluate if our proposed method can benefit skill learning for dynamic robotic tasks, we investigate if our model can adapt online to changing goals, which is a natural property of reinforcement learning policies as described in Sec.~\ref{sec:adapt}. In the first experiment, we let the STM synthesize a trajectory that makes the gripper reach to the goal position shown in blue in Fig.~\ref{fig:sawyer_changing_goals}. Midway through the execution, we change the goal coordinates $\psi_t$ and observe that our model is able to quickly adapt to this change and servo the gripper to the new goal location, exceeding the length of all demonstration trajectories our model was trained on. In our second experiment, Sawyer picks up the block from a preset location (drawn with black solid lines in Fig.~\ref{fig:sawyer_changing_goals}), as in the pick-and-place experiment in Sec.~\ref{sec:stacking}. After grasping, we change the goal location of the block (visualized by the solid orange box) and the STM exhibit fast adaption to these new conditions. The adapted trajectory (green line) is close to the movement planned directly for the new goal location (blue line).

In combination with a learned IDM, we show in Fig.~\ref{fig:reacher_tracking} that skills learned in the simulator can be directly transferred to working torque-control policies in the real world. Moreover, once an IDM has been trained, it can be reused for any different servoing goals. While the sampling complexity is improved over conventional model-free reinforcement learning algorithms, our results exhibit jerky and significantly less precise behaviors compared to executing the STM's predicted state in position control mode. Extending our framework to have the STM be trained end-to-end with an IDM in the loop remains an future research direction, as it would lead to precise and sample-efficient low-level control policies that are competitive with policies learned via reinforcement learning.

\section{Conclusion}
\label{sec:conclusion}
In this work, we present a recurrent neural network architecture and training procedure that enables the efficient generation of complex joint position trajectories. Our experiments have shown that our STM can generalize to unseen tasks and is able to learn the underlying task specification which enables it to follow high-level instructions. In combination with a learned inverse dynamics model, we have shown a fully trainable motion planning pipeline on a real robot that combines the state transition model, as planning module, with an IDM, as a joint-position controller, to generate joint torque commands that tracks the synthesized trajectories.

In our evaluation of the model in combination with an IDM, we have observed successful transfer of basic closed-loop control policies. However, in many cases, the IDM was not able to exactly reach the planned state by the STM, leading to tracking errors. While the STM was able to generate trajectories that far exceeded the length of the demonstrations, it ultimately has limits that prohibit it to adapt to any unseen task if the goal is moved to far outside the state space the model was trained on. It is a general problem in state-of-the-art machine learning models that predictions are bounded by the training data. Future work needs to tackle this fundamental issue, \eg{} by introducing an inductive bias to the model that improves its generalizability over a particular set of tasks.

We are excited about the range of potential future directions. Specifically, given the adaptability of our model, it is interesting to explore how it could generalize towards more complex tasks with human teammates. We would also like to deepen the connection of these models with trajectory optimization methods, which would allow for trajectories that optimize over
a variety of dynamic and task-based criteria.


\bibliographystyle{IEEEtran}
\footnotesize{
\bibliography{literature}
}

\end{document}